\documentclass[runningheads]{llncs}

\usepackage{graphicx}
\usepackage{amsmath}
\usepackage{amssymb}
\usepackage{xspace}
\usepackage{hyperref}

\usepackage[margin=1.25in]{geometry}

\newcommand{\moving}{m}
\newcommand{\moved}{\moving \circ \phi}
\newcommand{\fixed}{f}
\newcommand{\movingseg}{s_m}
\newcommand{\movedseg}{\movingseg \circ \phi}
\newcommand{\fixedseg}{s_f}
\newcommand{\Loss}{\mathcal{L}}
\newcommand{\optimal}{\lambda^*}
\newcommand{\sumpairs}{\sum_{\substack{\moving, \fixed \in \mathcal{D}}}}

\newcommand{\methodname}{HyperMorph\xspace}
\newcommand{\subpara}[1]{\vspace{0.2cm} \noindent \textbf{#1.}}

\begin{document}

\title{\methodname: Amortized Hyperparameter Learning for Image Registration}
\titlerunning{\methodname}

\author{Andrew Hoopes\inst{1} \and
Malte Hoffmann\inst{1,2} \and
Bruce Fischl\inst{1-3} \and \\
John Guttag\inst{3} \and
Adrian V. Dalca\inst{1-3}}

\authorrunning{A. Hoopes et al.}

\institute{
Martinos Center for Biomedical Imaging, MGH \and
Department of Radiology, Harvard Medical School \and
Computer Science and Artificial Intelligence Lab, MIT
}

\maketitle

\begin{abstract}
We present \methodname, a learning-based strategy for deformable image registration that removes the need to tune important registration hyperparameters during training. Classical registration methods solve an optimization problem to find a set of spatial correspondences between two images, while learning-based methods leverage a training dataset to learn a function that generates these correspondences. The quality of the results for both types of techniques depends greatly on the choice of hyperparameters. Unfortunately, hyperparameter tuning is time-consuming and typically involves training many separate models with various hyperparameter values, potentially leading to suboptimal results. To address this inefficiency, we introduce amortized hyperparameter learning for image registration, a novel strategy to \textit{learn} the effects of hyperparameters on deformation fields. The proposed framework learns a hypernetwork that takes in an input hyperparameter and modulates a registration network to produce the optimal deformation field for that hyperparameter value. In effect, this strategy trains a single, rich model that enables rapid, fine-grained discovery of hyperparameter values from a continuous interval at test-time. We demonstrate that this approach can be used to optimize multiple hyperparameters considerably faster than existing search strategies, leading to a reduced computational and human burden as well as increased flexibility. We also show several important benefits, including increased robustness to initialization and the ability to rapidly identify optimal hyperparameter values specific to a registration task, dataset, or even a single anatomical region, all without retraining the \methodname model. Our code is publicly available at~\url{http://voxelmorph.mit.edu}.


\end{abstract}


\section{Introduction}

Deformable image registration aims to find a set of dense correspondences that accurately align two images. Classical optimization-based techniques for image registration have been thoroughly studied, yielding mature mathematical frameworks and widely used software tools~\cite{ashburner2007,avants2008,beg2005,rueckert1999,vercauteren2009}. Learning-based registration methods employ image datasets to learn a function that rapidly computes the deformation field between image pairs~\cite{balakrishnan2019,rohe2017,sokooti2017,de2019,wu2015,yang2017}. These methods involve choosing registration hyperparameters that dramatically affect the quality of the estimated deformation field. Optimal hyperparameter values can differ substantially across image modality and anatomy, and even small changes can have a large impact on accuracy. Choosing appropriate hyperparameter values is therefore a crucial step in developing, evaluating, and deploying registration methods.

\begin{figure}[t]
  \centering
  \includegraphics[width=1\textwidth]{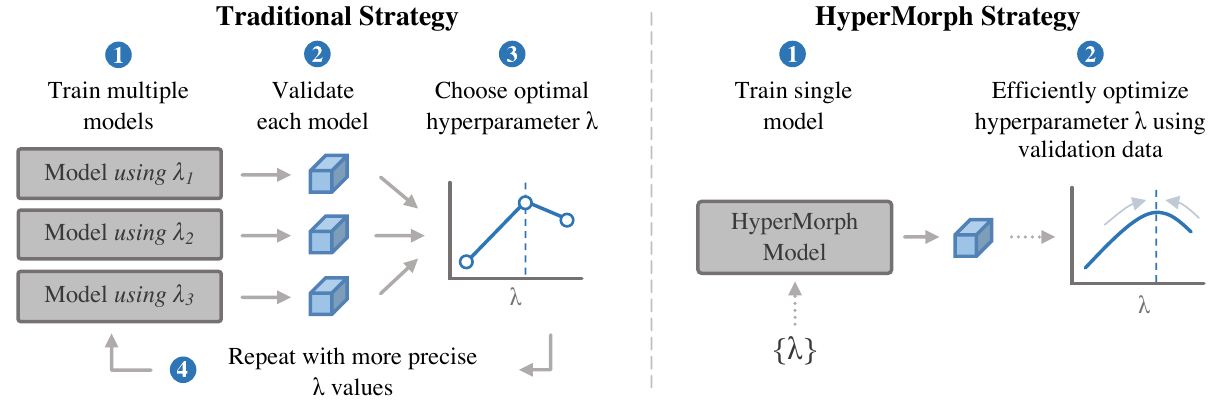}
  \caption{Hyperparameter optimization strategies. Traditional approaches (left) repeatedly train a registration model, each time with a different hyperparameter value. The proposed \methodname approach (right) optimizes a single, richer model once, which approximates a landscape of traditional models.}
  \label{fig:method-schematic}
\end{figure}

Tuning these hyperparameters most often involves grid or random search techniques to evaluate separate models for discrete hyperparameter values (Figure~\ref{fig:method-schematic}). In practice, researchers  typically perform a sequential process of optimizing and validating models with a small subset of hyperparameter values, adapting this subset, and repeating. Optimal hyperparameter values are selected based on model performance, generally determined by human evaluation or additional validation data such as anatomical annotations. This approach requires considerable computational and human effort, which may lead to suboptimal parameter choices, misleading negative results, and impeded progress, especially when researchers might resort to using values from the literature that are not adequate for their specific dataset or registration task.

In this work, we introduce a substantially different approach, \methodname, to tackle registration hyperparameters: amortized hyperparameter learning for image registration. Our contributions are:

\subpara{Method} We propose an end-to-end strategy to~\textit{learn} the effects of registration hyperparameters on deformation fields with a single, rich model, replacing the traditional hyperparameter tuning process (Figure~\ref{fig:method-schematic}). A \methodname model is a hypernetwork that approximates a landscape of registration networks for a range of hyperparameter values, by learning a continuous function of the hyperparameters. Users only need to learn a single \methodname model that enables rapid test-time image registration for any hyperparameter value. This eliminates the need to train a multitude of separate models each for a fixed hyperparameter, since \methodname accurately estimates their outputs at a fraction of the computational and human effort. In addition, \methodname enables rapid, accurate hyperparameter tuning for registration tasks involving many hyperparameters, in which computational complexity renders grid-search techniques ineffective.

\subpara{Properties} By exploiting implicit weight-sharing, a single \methodname model is efficient to train compared to training the many individual registration models it is able to encompass. We also show that \methodname is more robust to initialization than standard registration models, indicating that it better avoids local minima while reducing the need to retrain models with different initializations.

\subpara{Utility} \methodname enables rapid discovery of optimal hyperparameter values at \textit{test-time}, either through visual assessment or automatic optimization in the continuous hyperparameter space. We demonstrate the substantial utility of this approach by using a \textit{single} \methodname model to identify the optimum hyperparameter values for different datasets, different anatomical regions, or different registration tasks. \methodname also offers more precise tuning compared to grid or sequential search.


\section{Related Work}

\subpara{Image Registration}
Classical approaches independently estimate a deformation field by optimizing an energy function for each image pair. These include elastic models~\cite{bajcsy1989}, b-spline based deformations~\cite{rueckert1999}, discrete optimization methods~\cite{dalca2016,glocker2008},
Demons~\cite{vercauteren2009}, SPM~\cite{ashburner2000}, LDDMM~\cite{beg2005,cao2005,joshi2000,miller2005,zhang2017},
DARTEL~\cite{ashburner2007}, and symmetric normalization (SyN)~\cite{avants2008}.
Recent learning-based approaches make use of convolutional neural networks (CNNs) to learn a function that rapidly computes the deformation field for an image pair. Supervised models learn to reproduce deformation fields estimated or simulated by other methods~\cite{krebs2017,rohe2017,sokooti2017,yang2017}, whereas unsupervised strategies train networks that optimize a loss function similar to classical cost functions and do not require the ground-truth registrations needed by supervised methods~\cite{balakrishnan2019,dalca2019varreg,hoffmann2020,krebs2019,de2019}.

Generally, all these methods rely on at least one hyperparameter that balances the optimization of an image-matching term with that of a regularization or smoothness term. Additional hyperparameters are often used in the loss terms, such as the neighborhood size of local normalized cross-correlation~\cite{avants2011} or the number of bins in mutual information~\cite{viola1997alignment}. Choosing optimal hyperparameter values for classical registration algorithms is a tedious process since pair-wise registration typically requires tens of minutes or more to compute. While learning-based methods enable much faster test-time registration, individual model \textit{training} is expensive and can require days to converge, causing the hyperparameter search to consume hundreds of GPU-hours~\cite{balakrishnan2019,hoffmann2020,de2019}.

\subpara{Hyperparameter Optimization}
Hyperparameter optimization algorithms jointly solve a validation objective with respect to model hyperparameters and a training objective with respect to model weights~\cite{franceschi2018}. The simplest approach treats model training as a black-box function, including grid, random, and sequential search~\cite{bergstra2012}. Bayesian optimization is a more sample-efficient strategy, leveraging a probabilistic model of the objective function to search and evaluate hyperparameter performance~\cite{bergstra2011}. Both approaches are often inefficient, since the algorithms involve repeated optimizations for each hyperparameter evaluation.
Enhancements to these strategies have improved performance by extrapolating learning curves before full convergence~\cite{domhan2015,klein2016} and evaluating low-fidelity approximations of the black-box function~\cite{kandasamy2017}. Other adaptations use bandit-based approaches to selectively allocate resources to favorable models~\cite{jamieson2016,li2017}.
Gradient-based techniques differentiate through the nested optimization to approximate gradients as a function of the hyperparameters~\cite{luketina2016,maclaurin2015,pedregosa2016}. These approaches are computationally costly and require evaluation of a metric on a comprehensive, labeled validation set, which may not be available for every registration task.

\subpara{Hypernetworks}
Hypernetworks are networks that output weights of a primary network~\cite{ha2016,Klocek2019,schmidhuber1993}. Recently, they have gained traction as efficient methods of gradient-based hyperparameter optimization since they enable easy differentiation through the entire model with respect to the hyperparameters of interest. For example, SMASH uses hypernetworks to output the weights of a network conditioned on its architecture~\cite{brock2017}. Similar work employs hypernetworks to optimize weight decay in classification networks and demonstrates that sufficiently sized hypernetworks are capable of approximating its global effect~\cite{lorraine2018,mackay2019}. HyperMorph extends hypernetworks, combining them with learning-based registration to estimate the effect of hyperparameter values on deformations.


\section{Methods}

\subsection{\methodname}

Deformable image registration methods find a dense, non-linear correspondence field~$\phi$ between a moving image~$\moving$ and a fixed image~$\fixed$, and can employ a variety of hyperparameters. We follow current unsupervised learning-based registration methods and define a network~$g_{\theta_g}(\moving, \fixed) = \phi$ with parameters~$\theta_g$ that takes as input the image pair~$\{\moving, \fixed\}$ and outputs the optimal deformation field~$\phi$. 

\begin{figure}[t]
\centering
  \includegraphics[width=1\textwidth]{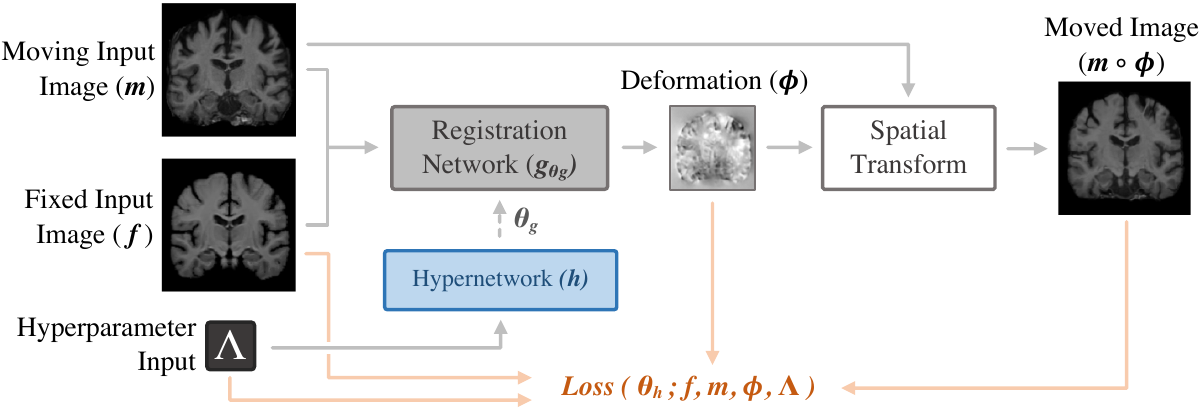}
  \caption{\methodname framework. A hypernetwork (blue) learns to output the parameters of a registration network given registration hyperparameters~$\Lambda$. \methodname is trained end-to-end, exploiting implicit weight-sharing among the full landscape of registration networks within a continuous interval of hyperparameter values.
  }
  \label{fig:architecture}
\end{figure}

Our key idea is to model a hypernetwork that learns the effect of \textit{loss} hyperparameters on the desired registration. Given loss hyperparameters~$\Lambda$ of interest, we define the hypernetwork function~$h_{\theta_h}(\Lambda) = \theta_g$ with parameters~$\theta_h$ that takes as input sample values for~$\Lambda$ and outputs the parameters of the registration network~$\theta_g$ (Figure \ref{fig:architecture}). We learn optimal hypernetwork parameters~$\theta_h$ using stochastic gradient methods, optimizing the loss
\begin{alignat}{3}
\Loss_h(\theta_h; \mathcal{D})
&=& \mathbb{E}_{\Lambda \sim p(\Lambda)} \Big[ \Loss(\theta_h ; \mathcal{D}, &\Lambda)& \Big] ,
\label{eq:hyper-loss}
\end{alignat}
where~$\mathcal{D}$ is a dataset of images,~$p(\Lambda)$ is a prior probability over the hyperparameters, and~$\Loss(\cdot)$ is a registration loss involving hyperparameters~$\Lambda$. For example, the distribution~$p(\Lambda)$ can be uniform over some predefined range, or it can be adapted based on prior expectations. At every mini-batch, we sample a set of hyperparameter values from this distribution and use these both as input to the network~$h_{\theta_h}(\cdot)$ and in the loss function~$\Loss(\cdot)$ for that iteration.

\subpara{Unsupervised Model Instantiations} Following unsupervised leaning-based registration, we use the loss function:
\begin{align}
\Loss_h(\theta_h; \mathcal{D})
= \mathbb{E}_\Lambda \Big[ 
\sumpairs \Big(
(1 - \lambda) \Loss_{sim}(\fixed, \moved; \lambda_{sim}) + \lambda \Loss_{reg}(\phi; \lambda_{reg}) \Big) \Big], \hfill \quad
\raisetag{2\normalbaselineskip}
\label{eq:hyper-loss-classical}
\end{align}
where~$\moved$ represents~$\moving$ warped by~$\phi=g_{\theta_g}(\moving, \fixed)$,~$\theta_g = h_{\theta_h}(\Lambda)$. The loss term~$\Loss_{sim}$ measures image similarity and might involve hyperparameters~$\lambda_{sim}$, whereas~$\Loss_{reg}$ quantifies the spatial regularity of the deformation field and might involve hyperparameters~$\lambda_{reg}$. The regularization hyperparameter~$\lambda$ balances the relative importance of the separate terms, and~$\Lambda = \{\lambda, \lambda_{sim}, \lambda_{reg}\}$.

When registering images of the same modality, we use standard similarity metrics for~$\Loss_{sim}$: mean-squared error (MSE) and \textit{local} normalized cross-correlation (NCC). Local NCC includes a hyperparameter defining the neighborhood size. For cross-modality registration, we use normalized mutual information (NMI), which involves a hyperparameter controlling the number of histogram bins~\cite{viola1997alignment}.

We parameterize the deformation field~$\phi$ with a stationary velocity field (SVF) $v$ and integrate it within the network to obtain a diffeomorphism, which is invertible by design~\cite{arsigny2006log,ashburner2007,dalca2019varreg}. We regularize $\phi$ using \mbox{$\Loss_{reg}(\phi) = \frac{1}{2}\|\nabla v\|^2$}.

\subpara{Semi-supervised Model Instantiation} Building on recent learning-based methods that use additional volume information during training~\cite{balakrishnan2019,hoffmann2020,hu2018weakly},
we also apply \methodname to the semi-supervised setting by modifying the loss function to incorporate existing training segmentation maps:
\begin{alignat}{3}
\Loss_h(\theta_h; \mathcal{D}) = \mathbb{E}_\Lambda 
\sumpairs  &\Big[ 
(1-\lambda)(1-\gamma) \Loss_{sim}(\fixed, \moved; \lambda_{sim})  \Big. \nonumber\\[-10pt] 
& \Big. + \lambda \Loss_{reg}(\phi; \lambda_{reg}) + (1 - \lambda) \gamma \Loss_{seg}(\fixedseg, \movedseg)  \Big] ,
\label{eq:hyper-loss-semisupervised}
\end{alignat}
where~$\Loss_{seg}$ is a segmentation similarity metric, usually the Dice coefficient~\cite{dice1945},
weighted by the hyperparameter~$\gamma$, and~$\movingseg$ and~$\fixedseg$ are the segmentation maps of the moving and fixed images, respectively.

\subsection{Hyperparameter Tuning}

Given a test image pair~$\{\moving, \fixed\}$, a trained \methodname model can efficiently yield the deformation field as a function of important hyperparameters. If no external information is available, optimal hyperparameters may be rapidly tuned in an interactive fashion. However, landmarks or segmentation maps are sometimes available for validation subjects, enabling rapid automatic tuning.

\subpara{Interactive} Sliders can be used to change hyperparameter values in near real-time until the user is visually satisfied with the registration of some image pair~$\{\moving, \fixed\}$. In some cases, the user might choose different settings when studying specific regions of the image. For example, the optimal value of the~$\lambda$ hyperparameter (balancing the regularization and the image-matching term) can vary by anatomical structure in the brain (see Figure~\ref{fig:opt-labels}). This interactive tuning technique is possible because of  the \methodname ability  to efficiently yield the effect of~$\lambda$ values on the deformation~$\phi$.

\subpara{Automatic} If segmentation maps~$\{\movingseg,\fixedseg\}$ are available for validation, a single trained \methodname model enables hyperparameter optimization using
\begin{align}
\Lambda^* = \arg\max_\Lambda &\Loss(\Lambda; \theta_h, \mathcal{D}, \mathcal{V}) 
= \arg\max_\Lambda
\sum_{\substack{(\moving,\fixed) \in \mathcal{D}^2 \\ (\movingseg, \fixedseg) \in \mathcal{V}^2}}
\Loss_{val}(\fixedseg, \movingseg \circ \phi), 
\label{eq:val-hyper-loss}
\end{align}
where~$\mathcal{V}$ is a set of validation segmentation maps and $\phi = g_{h(\Lambda)}(\moving, \fixed)$, as before. We implement this optimization by freezing the learned hypernetwork parameters~$\theta_h$, treating the input~$\Lambda$ as a parameter to be learned, and using stochastic gradient strategies to rapidly optimize~\eqref{eq:val-hyper-loss}.

\subsection{Implementation}

The hypernetwork we use in the experiments consists of four fully connected layers, each with 64 units and ReLu activation except for the final layer, which uses Tanh activations. The proposed method applies to any registration network architecture, and we treat the hypernetwork and the registration network as a single, large network. The only trainable parameters~$\theta_h$ are those of the hypernetwork. We implement \methodname with the open-source VoxelMorph library~\cite{balakrishnan2019}, using a U-Net-like~\cite{ronneberger2015} registration architecture. The U-Net in this network consists of a 4-layer convolutional encoder (with 16, 32, 32, and 32 channels), a 4-layer convolutional decoder (with 32 channels for each layer), and 3 more convolutional layers (of 32, 16, and 16 channels). This results in a total of 313,507 convolutional parameters that are provided as output by the hypernetwork. We use the ADAM optimizer~\cite{kingma2014adam} during training.


\section{Experiments}

We demonstrate that a single \methodname model performs on par with and captures the behavior of a rich landscape of individual registration networks trained with separate hyperparameter values, while incurring substantially less computational cost and human effort. We test models with one or two registration hyperparameters. Next, we illustrate considerable improvements in robustness to initialization. Finally, we demonstrate the powerful utility of \methodname for rapid hyperparameter optimization at validation --- for different subpopulations of data, registration types, and individual anatomical structures.

\subpara{Datasets}
We use two large sets of 3D brain magnetic resonance (MR) images. The first is a multi-site dataset of 30,495 T1-weighted (T1w) scans gathered across 8 public datasets:
ABIDE~\cite{di2014},
ADHD200~\cite{milham2012},
ADNI~\cite{adni2003},
GSP~\cite{dagley2017},
MCIC~\cite{gollub2013},
PPMI~\cite{marek2011},
OASIS~\cite{marcus2007},
and UK Biobank~\cite{sudlow2015}.
We divide this dataset into train, validation, and test sets of sizes 10,000, 10,000, and 10,495, respectively. The second dataset involves a multi-modal collection of 1,558 T1w, T2-weighted (T2w), multi-flip-angle, and multi-inversion-time images gathered from in-house data and the public ADNI and HCP~\cite{bookheimer2019} datasets. We divide this dataset into train, validation, and test sets of sizes 528, 515, and 515, respectively. All MRI scans are conformed to a 256$\times$256$\times$256 1-mm isotropic grid space, bias-corrected, and skull-stripped using FreeSurfer~\cite{fischl2012}, and we also produce automated segmentation maps for evaluation. We affinely normalize and uniformly crop all images to 160$\times$192$\times$224 volumes.

\subpara{Evaluation}
For evaluation, we use the volume overlap of anatomical label maps using the Dice metric~\cite{dice1945}.

\subpara{Baseline Models}
\methodname can be applied to any learning-based registration architecture, and we seek to validate its ability to capture the effects of hyperparameters on the inner registration network~$g_{\theta_g}(\cdot)$. To enable this insight, we train standard VoxelMorph models with architectures identical to~$g_{\theta_g}(\cdot)$ as baselines, each with its fixed set of hyperparameters.

\subsection{Experiment 1: \methodname Efficiency and Capacity}

\begin{figure}[t]
  \centering
  \includegraphics[width=\textwidth]{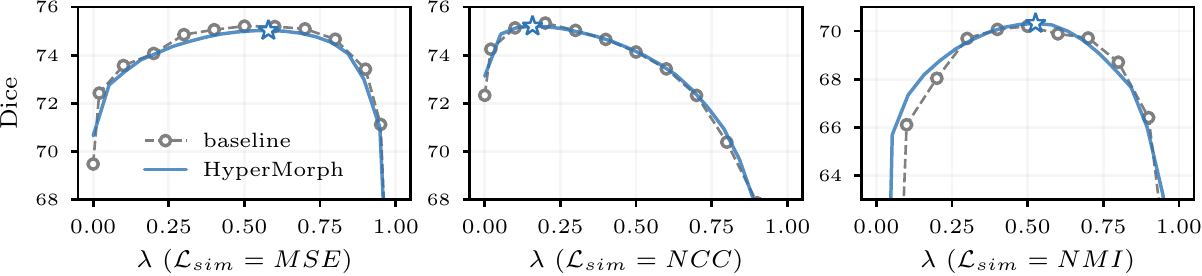}
  \caption{Mean Dice scores achieved by a single \methodname model (blue) and baselines trained for different regularization weights~$\lambda$ (gray) when using each of the MSE, NCC and NMI similarity metrics, respectively. Optima~$\optimal$ computed with \methodname are indicated by the star markers.
  }
  \label{fig:baseline-comparison}
\end{figure}

We aim to evaluate if a single \methodname is capable of encapsulating a landscape of baseline models.

\subpara{Setup}
We first assess how the accuracy and computational cost of a single \methodname model compare to standard grid hyperparameter search for the regularization weight~$\lambda$.
We separately train \methodname as well as VoxelMorph baselines using the similarity metrics MSE (scaled by a constant estimated image noise) and NCC (with window size $9^3$) for within-modality registration and NMI (with 32 fixed bins) for cross-modality registration, for which we train 13, 13, and 11 baseline models, respectively. We validate the trained networks on 100 random image pairs for visualization. For hyperparameter optimization after training, we use a subset of 20 pairs.

Additionally, we assess the ability of \methodname to learn the effect of multiple hyperparameters simultaneously. We first train a \methodname model treating $\lambda$ and the local NCC window size as hyperparameters. We also train a semi-supervised \methodname model based on a subset of six labels, and hold out six other labels for validation. In this experiment, the hyperparameters of interest are~$\lambda$ and the relative weight~$\gamma$ of the semi-supervised loss~\eqref{eq:hyper-loss-semisupervised}. Training baselines requires a two-dimensional grid search on 3D models and is computationally prohibitive. Consequently, we conduct these experiments in 2D on a mid-coronal slice, using baselines for 25 combinations of hyperparameter values.

\subpara{Results} 
\textit{Computational Cost}. A single \methodname model requires substantially less time to convergence than a baseline-model grid search. For single-hyperparameter tests, \methodname requires $5.2~\pm 0.2$ times fewer GPU-hours than a grid search with baseline models (Table~\ref{tab:dice-runtime-table}). For models with two hyperparameters, the difference is even more striking, with \methodname requiring $10.5 \pm 0.2$ times fewer GPU-hours than the baseline models. 

\begin{figure}[t]
  \centering
  \includegraphics[width=\textwidth]{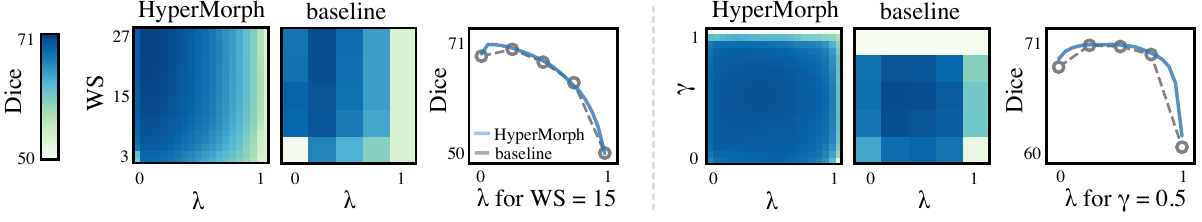}
  \caption{Two-dimensional hyperparameter search. Left: unsupervised registration with regularization weight~$\lambda$ and local NCC window size WS. Right: semi-supervised registration with hyperparameters~$\lambda$ and segmentation supervision weight~$\gamma$. For the semi-supervised models, we compute total Dice on both training and held-out labels.
  }
  \label{fig:multi-hyp-baseline-comparison}
\end{figure}

\textit{Performance}. Figures \ref{fig:baseline-comparison} and \ref{fig:multi-hyp-baseline-comparison} show that \methodname yields optimal hyperparameter values similar to those obtained from a dense grid of baseline models despite the significant computational advantage. An average difference in the optimal hyperparameter value~$\optimal$ of only $0.04 \pm 0.02$ across single-hyperparameter experiments results in a negligible maximum Dice difference of $0.16 \pm 0.03$ (on a scale of $0$ to $100$). Similarly, multi-hyperparameter experiments yield a maximum Dice difference of only $0.32 \pm 0.02$. In practice, fewer baselines might be trained at first for a coarser hyperparameter search, resulting in either suboptimal hyperparameter choice or sequential search leading to substantial manual overhead. 

Overall, a \textit{single} \methodname model is able to capture the behavior of a range of baseline models individually optimized for different hyperparameters, facilitating optimal hyperparameter choices and accuracy at a substantial reduction in computational cost. We emphasize that the goal of the experiment is not to compare \methodname to a particular registration tool, but to demonstrate the effect that this strategy can have on an existing registration network.

\begin{table}[b]
\centering
\caption{Comparison between HyperMorph and baseline grid-search techniques for model variability across random initializations (left) and for runtimes (right). We provide runtimes separately for experiments with 1 and 2 hyperparameters (HP).}
\begin{tabular}{|l|cc|cc|}
\hline
 & \multicolumn{2}{c|}{Robustness (init SD)} & \multicolumn{2}{c|}{Runtime (total GPU-hours)} \\
  & MSE & NMI & \begin{tabular}[c]{@{}c@{}}1 HP (3D)\end{tabular} & \begin{tabular}[c]{@{}c@{}}2 HPs (2D)\end{tabular} \\ \hline
HyperMorph & \textbf{1.97e-1} & \textbf{2.46-1} & \textbf{146.9~$\pm$ 32.0}  & \textbf{4.2~$\pm$ 0.6} \\ 
Baseline & 5.50e-1 & 5.32e-1 & 765.3~$\pm$ 249.1 & 44.0~$\pm$ 4.6 \\ \hline
\end{tabular}
\label{tab:dice-runtime-table}
\end{table}

\subsection{Experiment 2: Robustness to Initialization}

\subpara{Setup} We evaluate the robustness of each strategy to network initialization. We repeat the previous, single-hyperparameter experiment with MSE and NMI, retraining four \methodname models and four \textit{sets} of baselines each trained for five values of hyperparameter~$\lambda$. For each training run, we re-initialize all kernel weights using Glorot uniform~\cite{glorot2010understanding} with a different seed. We evaluate each model using 100 image pairs and compare the standard deviation~(SD) across initializations of the \methodname and baseline networks. 

\begin{figure}[t]
  \centering
  \includegraphics[width=0.8\textwidth]{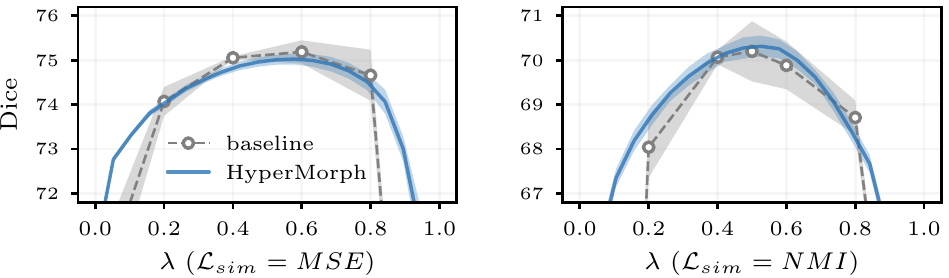}
  \caption{Variability across several model initializations for \methodname and baselines. The shaded areas indicate the standard deviation of registration accuracy, which is substantially more narrow for \methodname.
  }
  \label{fig:robustness}
\end{figure}

\subpara{Results} Figure~\ref{fig:robustness} shows that \methodname is substantially more robust (lower SD) to initialization compared to the baselines, suggesting that \methodname is less likely to converge to local minima. Across the entire range of~$\lambda$, the average Dice SD for HyperMorph models trained with MSE is~$2.79$ times lower than for baseline SD, and for NMI-trained models, \methodname SD is~$2.16$ times lower than baseline SD (Table \ref{tab:dice-runtime-table}). This result further emphasizes the computational efficiency provided by \methodname, since in typical hyperparameter searches, models are often trained multiple times for each hyperparameter value to negate potential bias from initialization variability.

\subsection{Experiment 3: Hyperparameter-Tuning Utility}

\subpara{Setup}
\textit{Interactive Tuning}. We demonstrate the utility of \methodname through an interactive tool that enables visual optimization of hyperparameters even if no segmentation data are available. The user can explore the effect of \textit{continuously varying} hyperparameter values using a single trained model and choose an optimal deformation manually at high precision. Interactive tuning can be explored at~\url{http://voxelmorph.mit.edu}.

\textit{Automatic Tuning}. When anatomical annotations are available for validation, we demonstrate rapid, automatic optimization of the hyperparameter~$\lambda$ across a variety of applications. In each experiment, we identify the optimal regularization weight~$\optimal$ given 20 registration pairs and use 100 registration pairs for evaluation. First, we investigate how $\optimal$ differs across subpopulations and anatomical regions. We train \methodname on a subset of image pairs across the entire T1w training set, and at validation we optimize~$\lambda$ separately for each of ABIDE, GSP, PPMI, and UK Biobank. With this same model, we identify~$\optimal$ separately for each of 10 anatomical regions. Second, we explore how~$\optimal$ differs between cross-sectional and longitudinal registration; for \methodname trained on both within-subject and cross-subject pairs from ADNI, we optimize~$\lambda$ separately for validation pairs within and across subjects.

\begin{figure}[t]
  \centering
  \includegraphics[width=\textwidth]{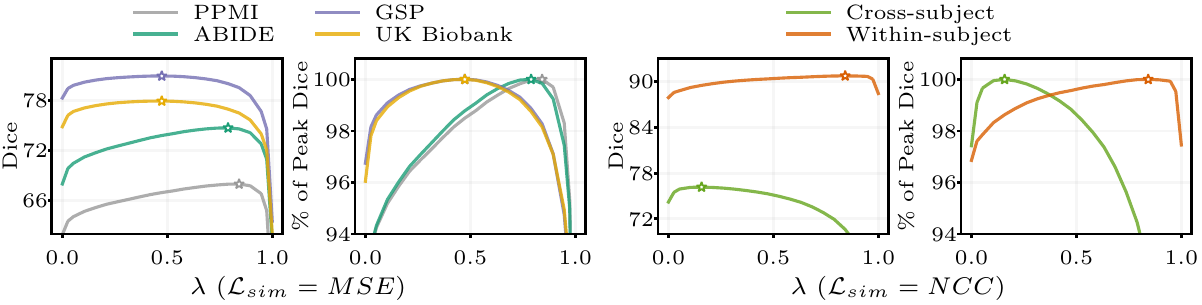}
  \caption{Registration accuracy across dataset subpopulations (left) and registration tasks (right). The stars indicate the optimal value~$\optimal$ as identified by automatic hyperparameter optimization.
  }
  \label{fig:subtask-optimization}
\end{figure}

\begin{figure}[b]
  \centering
  \includegraphics[width=\textwidth]{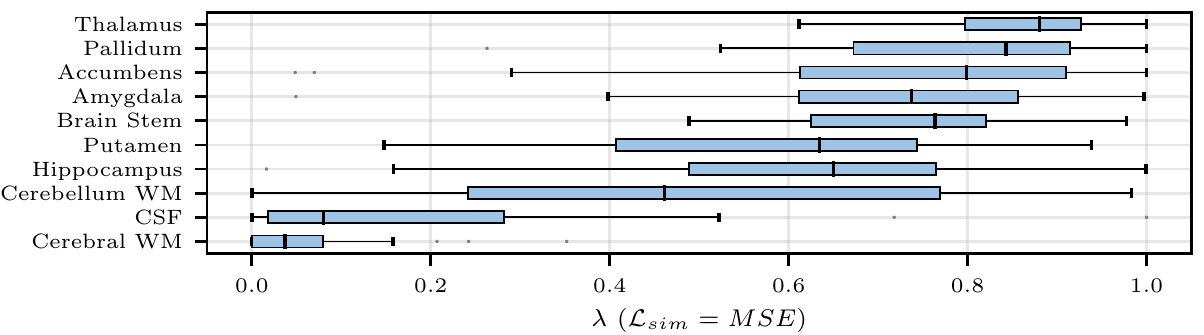}
  \caption{Optimal regularization weights~$\optimal$ across individual anatomical labels. The ranges shown are estimated with \methodname for each set of image pairs.
  }
  \label{fig:opt-labels}
\end{figure}

\subpara{Results}
Figures~\ref{fig:subtask-optimization} and~\ref{fig:opt-labels} show that $\optimal$ varies substantially across subpopulations, registration tasks, and anatomical regions. For example, PPMI and ABIDE require a significantly different value of $\optimal$ than GSP and the UK Biobank. Importantly, with a suboptimal choice of hyperparameters, these datasets would have yielded considerably lower registration quality (Dice scores). The variability in the optimal hyperparameter values is likely caused by differences between the datasets; the average age of the ABIDE population is lower than those of other datasets, while the PPMI scans are of lower quality. Similarly, cross-subject and within-subject registration require different levels of regularization. Finally, Figure~\ref{fig:opt-labels} illustrates that $\optimal$ varies by anatomical region, suggesting that regularization weights should be chosen by users depending on their tasks downstream from the registration. On average, the automatic hyperparameter optimization takes just~$2.8 \pm 0.3$ minutes using 20 validation pairs.

The vast majority of existing registration pipelines assume a single hyperparameter value to be optimal for an entire dataset, or even across multiple datasets. Our results highlight the importance of \methodname as a rapid, easy-to-use tool for finding optimal hyperparameters, interactively or automatically, for different subpopulations, tasks, or even individual anatomical regions, without the need to retrain models.

\section{Conclusion}

The accuracy of deformable image registration algorithms greatly depends upon the choice of hyperparameters. In this work, we present \methodname, a learning-based strategy that removes the need to repeatedly train models to quantify the effects of hyperparameters on model performance. \methodname employs a hypernetwork which takes the desired hyperparameters as input and predicts the parameters of a registration network tuned to these values. In contrast to existing learning-based methods, \methodname estimates optimal deformation fields for arbitrary image pairs and \textit{any} hyperparameter value from a continuous interval by exploiting sharing of similar weights across the landscape of registration networks. A single \methodname model then  enables fast hyperparameter tuning at test-time, requiring dramatically less compute and human time compared to existing methods. This is a significant advantage over registration frameworks that are optimized across discrete, predefined hyperparameter values in the hope of finding an optimal configuration.

We demonstrate that a single \methodname model facilitates discovery of optimal hyperparameter values for different dataset subpopulations, registration tasks, or even individual anatomical regions. This last result indicates a potential benefit and future direction of estimating a spatially varying field of smoothness hyperparameters for simultaneously optimal registration of all anatomical structures.
\methodname also provides the flexibility to identify the ideal hyperparameter for an individual image pair. For example, a pair of subjects with very different anatomies would benefit from weak regularization allowing warps of high non-linearity. We are eager to explore hypernetworks for a greater number of hyperparameters. We believe \methodname will drastically alleviate the burden of retraining networks with different hyperparameter values, thereby enabling efficient development of finely optimized models for image registration.

\section*{Acknowledgements}

Support for this research was provided by BRAIN U01 MH117023, NIBIB P41 EB015896, R01 EB023281, R01 EB006758, R21 EB018907, R01 EB019956, P41 EB030006, NIA R56 AG064027, R01 AG064027, AG008122, AG016495, NIMH R01 MH123195, NINDS R01 NS0525851, R21 NS072652, R01 NS070963 and NS083534, U01 NS086625, U24 NS10059103, R01 NS105820, NIH BNR U01 MH093765, the HCP, NICHD K99 HD101553, SIG RR023401, RR019307, and RR023043, and the Wistron Corporation. BF has a financial interest in CorticoMetrics which is reviewed and managed by MGH and MGB.

\bibliographystyle{splncs04}
\bibliography{references}

\end{document}